# A Scalable Predictive Modelling Approach to Identifying Duplicate Adverse Event Reports for Drugs and Vaccines

Jim W. Barrett[1], Nils Erlanson[1], Joana Félix China[1], G. Niklas Norén[1]

[1]Uppsala Monitoring Centre. Uppsala, Sweden.

Correspondence to jim.barrett@who-umc.org

## Abstract

### Objectives
To advance state-of-the-art for duplicate detection in large-scale pharmacovigilance databases and achieve more consistent performance across adverse event reports from different countries.

### Background
Unlinked adverse event reports referring to the same case impede statistical analysis and may mislead clinical assessment. Pharmacovigilance relies on large databases of adverse event reports to discover potential new causal associations, and computational methods are required to identify duplicates at scale. Current state-of-the-art is statistical record linkage which outperforms rule-based approaches. In particular, *vigiMatch* is in routine use for VigiBase, the WHO global database of adverse event reports, and represents the first statistical duplicate detection approach in pharmacovigilance deployed at scale. Originally developed for both medicines and vaccines, its application to vaccines has been limited due to inconsistent performance across countries.

### Methods
This paper extends vigiMatch from probabilistic record linkage to predictive modelling, refining features for medicines, vaccines, and adverse events using country-specific reporting rates, extracting dates from free text, and training separate support vector machine classifiers for medicines and vaccines. Recall was evaluated using 5 independent labelled test sets. Precision was assessed by annotating random selections of report pairs classified as duplicates.

### Results
Precision for the new method was 92% for vaccines and 54% for medicines, compared with 41% for the comparator method. Recall ranged from 80-85% across test sets for vaccines and from 40–86% for medicines, compared with 24–53% for the comparator method.

### Conclusion
Predictive modeling, use of free text, and country-specific features advance state-of-the-art for duplicate detection in pharmacovigilance.

## 1 Introduction
Pharmacovigilance is the science and activities relating to the detection, assessment, understanding and prevention of adverse effects or any other medicine/vaccine related problem (1). Randomized controlled trials are performed prior to regulatory approval to establish the efficacy and basic safety of new medicines and vaccines for human use. However, these trials are usually carried out in tightly controlled settings which may not fully reflect real-world conditions. They are not large enough to detect very rare adverse reactions and often exclude vulnerable patients. Therefore, the safety and efficacy of medicines need to be continually re-evaluated throughout their lifecycle.

Most countries have systems set up to collect individual case reports of adverse events possibly associated with medicines. These are monitored for information suggestive of causal associations between medicines and adverse events or new aspects of known associations, generally referred to as signals of suspected causality (2) and they remain the main source of post-marketing signals (3–5). VigiBase, the WHO global database of adverse event reports for medicines and vaccines, brings together 44 million reports from member organizations across 160 countries who participate in the WHO Programme for International Drug Monitoring (January 2026).

Individual case report systems have inherent limitations, including the largely voluntary nature of reporting. This can affect both reporting rates and data quality, thereby requiring clinical verification of signals before further action. A significant data quality challenge is the duplication of reports. Duplicates are unlinked reports that describe the same case of an adverse event for a specific patient at a certain time (6). Duplication may result from a failure to link follow-up reports with earlier reports, different reporters submitting reports for the same case, or the same reporter submitting reports regarding the same case to multiple pharmacovigilance organizations. Replication of the same case across databases of multiple organizations is another source of duplication (7). This is particularly problematic for reports derived from scientific publications, which are screened by many pharmacovigilance organizations. Duplication can be an obstacle to analysis of individual case reports, affecting both regulatory authorities and pharmaceutical companies (8,6,9–13). Pharmacovigilance organizations may combine expert human review with computational and statistical methods in signal detection and analysis, each of which can be misled by duplication. In statistical signal detection, false positives due to inflated counts may distract from more important case series. In expert review of collections of case reports related to a specific medicine or adverse event, duplicates may impede review and lead to incorrect conclusions.

To handle duplicates, they must first be detected and managed, which ideally should occur during the reporting or case processing stages. However, this is not always possible, and research has focused on duplicate detection prior to or in association with analysis (6,8–10,14). However, reports do not always contain enough information to confidently conclude that a given pair are duplicates. Conversely, very similar reports are not necessarily duplicates (15), especially for vaccines where campaigns targeting homogenous patient groups can lead to many similar reports. These problems are exacerbated in settings where data sharing constraints lead to reports with very limited information.

vigiMatch (8), hereafter referred to as vigiMatch2017, has been in continuous operational use for over a decade in VigiBase, and represents the first statistical duplicate detection approach in pharmacovigilance deployed at scale. A performance evaluation within the IMI PROTECT project found vigiMatch2017 to outperform rule-based methods, identifying duplicates in VigiBase that had not been detected by rule-based methods at national regulatory authorities, and some of those which were detected had been misclassified by human assessors overwhelmed by large numbers of false positives (6). Its real-world deployment has highlighted opportunities for improvement of the method. Specifically, vigiMatch2017's use of global reporting patterns in its probabilistic record linkage does not generalize well to reports from countries whose drug or adverse event profiles deviate substantially from the global averages, especially in the context of large public health programmes and mass administration campaigns, where patients, adverse events and dates of administration tend to have less variance, leading to high false positive rates for some countries.

In this study, we aim to develop and evaluate a duplicate detection method for large collections of individual case reports with improved precision and recall for medicines and vaccines, with consistent results across different settings. Through this paper, we will refer to the new method as vigiMatch2025.

# 2 Related Work

A recent review of duplicate detection methods in pharmacovigilance databases identified 25 scientific papers, most of which did not detail the methods used or their performance (13). Earlier publications have noted that many methods implemented in commercial and bespoke software are based on rules of various complexity including exact matching (6,12).

The vigiMatch2017 method for duplicate detection for de-identified reports in VigiBase and a similar method by the US FDA for the FDA adverse event reporting system (FAERS) have been described and evaluated in scientific publications (6,8–10,14). Both utilize probabilistic record linkage according to the principles outlined by Fellegi and Sunter (16) to detect duplicates even with mismatching details. At the same time, they require sufficient matching information, even if reports are identical. Some reflections on the two methods' similarities and relative strengths have previously been published (17). We consider them state-of-the-art for duplicate detection in collections of adverse event reports.

# 3 Data and Methods

## 3.1 VigiBase

VigiBase is the WHO global database of adverse event reports for medicines and vaccines. It is the largest collection of such reports in the world. For this study, we used a fixed version of VigiBase from 2$^{nd}$ January 2023, at which time the database comprised approximately 36 million reports. Adverse event reports include information related to the patient, the adverse event(s) and the medicinal product(s) which can be either in structured format or free text. Adverse events are coded using MedDRA® (the Medical Dictionary for Regulatory Activities terminology), which is the international medical terminology developed under the auspices of the International Council for Harmonisation of Technical Requirements for Pharmaceuticals for Human Use (ICH) (18). Drugs are coded using the WHODrug Global international reference drug dictionary for medicinal product information (19) with mappings to the Anatomical Therapeutic Chemical (ATC) classification system. VigiBase stores the adverse event reports following the ICH E2B(R2) specification (20).

We defined vaccine reports as any report involving a medicinal product in the J07 ATC class characterized in the report as either 'suspected' or 'interacting' (as opposed to 'concomitant'). We defined a drug report as any report which was not a vaccine report. We defined a pair of reports as a "vaccine pair" if it involved at least one vaccine report. Due to their overrepresentation and atypical reporting patterns, reports related to COVID-19 vaccines were excluded, leaving a dataset for this study with approximately 27 million drug reports and 1.8 million vaccine reports.

## 3.2 Reference datasets

Obtaining representative sets of duplicates for training and evaluating a duplicate detection method through random sampling is intractable, due to the extremely low prevalence of duplicates among all possible report pairs. For this study, we relied on an approach inspired by active learning to train our model and utilized a range of different reference sets to evaluate its recall. Evaluating the precision required a prospective random sampling of report pairs from the overall dataset. A summary of all labelled data used in this study is in table 1. Throughout this study, where labels did not already exist, report pairs were labelled as duplicates or non-duplicates following the annotation principles described in the supplementary materials (S1).

### 3.2.1 UMC labelled pairs

A set of 282 known duplicate pairs has historically been used to train vigiMatch2017, which includes the original 38 duplicate groups from its original publication (8). Further to these pairs, 1535 report pairs were labelled during development of vigiMatch2025.

Some of these were found by prospectively applying an existing model to extend the reference set (either vigiMatch2017 or an earlier version of vigiMatch2025). Others were found during preliminary evaluations of the model, or through debugging its implementation. All pairs were manually examined and annotated by at least one co-author.

These data were randomly split into training, validation and test sets with a ratio of 60:30:10 respectively. The test set was not accessed until the model was finalised, for the performance evaluations presented in this paper.

### 3.2.2 Externally sourced reference data

Four members of the WHO programme for international drug monitoring provided lists of confirmed duplicates from their own databases of adverse event reports. A subset of these could be retrieved in VigiBase and were included as reference sets for validation and evaluation of our method.

US Food and Drug Administration (FDA) Center for Drug Evaluation and Research (CDER) provided two sets of confirmed duplicates. *FDA_Gold* contained duplicates identified within 12 case series (2300 reports in total) in the FDA Adverse Event Reporting System (FAERS), doubly annotated by FDA safety reviewers, with adjudication by a third reviewer where necessary. *FDA_Silver* contained duplicates identified within 26 case series in FAERS (over 10,000 reports in total) by the FDA's duplicate detection system and validated by a single reviewer. From these datasets, 131 duplicate pairs in FDA_Gold, and 555 in FDA_Silver, could be linked to pairs of reports in the version of VigiBase used in this study. Linking the provided FAERS IDs to VigiBase is challenging, due to other national databases using similar ID schemes and some pairs may be linked erroneously. However, since this would lead to conservative estimates of recall, we decided not to manually verify all pairs.

With the assistance of the Spanish national regulatory authority AEMPS, Lareb Netherlands pharmacovigilance centre and the UK MHRA, we retrieved approximately 10,000 drug duplicate pairs and 325 vaccine duplicate pairs, either with explicitly supplied lists of confirmed duplicates or through linking reports using the E2B(R2) field A.1.11.2, which contains the case identifier from earlier transmissions (20). By construction, these reports will have an overrepresentation of the "externally indicated" feature described below, which we account for in our evaluation.

These datasets were not used for training. Pairs identified in the UMC dataset were excluded from these test sets. 50% of each were used for validation, and 50% held back for the final evaluations in the test set.

| **Source** | Number of Drug Pairs (duplicates/non-duplicates) | Number of Vaccine Pairs (duplicates/non-duplicates) | Labelling Criteria | % Used for Training | % Used for Validation | % Used for Testing |
|---|---|---|---|---|---|---|
| UMC | 630/498 | 316/373 | Single Annotator | 60 | 30 | 10 |
| FDA Gold | 131/0 | 0/0 | 2+ Annotators | 0 | 50 | 50 |
| FDA Silver | 555/0 | 0/0 | Single Annotator | 0 | 50 | 50 |
| LAREB | 1121/0 | 104/0 | Derived from e2b field A.1.11.2 with help from regulator | 0 | 50 | 50 |
| AEMPS | 1732/0 | 38/0 | | 0 | 50 | 50 |
| MHRA | 7260/0 | 183/0 | | 0 | 50 | 50 |

*Table 1: A summary of the labelled datasets used for training and evaluating the models in this study*

## 3.3 Duplicate detection method

### 3.3.1 Model overview

vigiMatch2025 computes numerical features designed to capture the similarity or dissimilarity between two reports and combines these into a composite match score for the pair using linear Support Vector Machine (SVM) predictive models. SVMs were chosen because of their flexibility, ability to perform well with a moderate amount of training data, the explainability of their predictions, and their greater on training examples close to the decision boundary. Separate SVM models were trained emphasis for drug pairs and vaccine pairs. Features for the predictive models were identified based on a subset of features from vigiMatch2017 complemented by new or modified features.

Following is a short summary of all features used in vigiMatch2025, with detailed definitions in the following sections.

New features in vigiMatch2025**:**

- **Externally indicated** (binary variable)
- **Date embeddings** (cosine similarity between convolved binary vectors)

Features in vigiMatch2025 inherited (*) or modified (†) from the vigiMatch2017 features:

- **Patient sex*** (hit-miss model for categorical variable)
- **Patient age at onset*** (hit-miss mixture model for a numerical variable)
- Reported **drugs†** and **adverse events†** (aggregated into a single feature through summation)
    - Reported drugs (country-specific hit-miss model for binary vector)
    - Reported adverse events (country-specific hit-miss model for binary vector)
    - Compensation for correlations between drug pairs, adverse event pairs, and drug-adverse event pairs (country-specific hit-miss model correlation compensation)
- **Adverse event onset date*** (hit-miss mixture model for the earliest adverse event onset date listed on the report)

vigiMatch2017 features not included in vigiMatch2025:

- **Patient initials** – Excluded because they are no longer available in VigiBase
- **Reporting country** – Excluded since the hit-miss model gives higher scores for matching on countries with fewer reports, effectively setting a lower threshold for suspected duplicates in those countries.

- **Outcome** – Excluded since information on outcome cannot be expected to be the same for e.g. unlinked follow-up reports.

A simple heuristic is also included in the model to exclude pairs which had overall mismatching demographic information, to avoid duplicate classifications being entirely driven by matches on drugs and adverse events. According to this heuristic, pairs were classified as non-duplicates if they had a negative net contribution from the patient age, patient sex and date embeddings features collectively.

Additionally, vigiMatch2025 incorporates a 'blocking' heuristic where it only ever considers report pairs that match on at least one drug (at the WHODrug active substance level) and include adverse events from at least one shared MedDRA System Organ Class. This is the same blocking scheme that is used in vigiMatch2017.

The SVMs were trained based on positive and negative controls from the UMC reference training dataset. To better reflect the class imbalance between duplicates and non-duplicates, and to ensure a diversity of negative controls amongst training samples, we complemented these negative controls with a random sample of $10^8$ pairs. We retained pairs with at least one shared block, resulting in approximately 2.4 million negative controls for the drug model and 6.2 million negative controls for the vaccine model. Due to the extremely low prevalence of duplicates amongst random pairs, this sample is unlikely to contain more than a few, if any, true duplicates.

### 3.3.2 Support Vector Machine

Support vector machines (SVMs) are a class of supervised machine learning model. Fundamentally, they find the widest boundary separating the positive and negative classes, whilst minimizing the number of misclassified examples from the training set. In contrast to many other supervised learning methods, inference is driven entirely by the training examples closest to the decision boundary (the so-called support vectors).

A weighted combination of these support vectors, together with an additional constant vector, defines a hyperplane which is the decision boundary. The sign of the distance between a new datapoint and this plane decides its classification. We chose to fit the model with a linear kernel, so that the decision function for predicting the label $y_i$ for feature vector $\vec{x}_i$ can be written:

$$y_i = \vec{w} \cdot \vec{x}_i + b$$

Where $\vec{w}$ and $b$ are the weights and intercept of the fitted support vector machine.

We used regularization penalty factor C=1, utilizing the implementation in the sklearn Python library (version 1.3.0) (21), which is in turn a wrapper around the libsvm library (22).

### 3.3.3 Hit-miss models

Several of the features in vigiMatch2025 are based on so-called hit-miss models which provide a mechanism to compute the log-likelihood ratio for a specific matching event (e.g. both reports are for 5-year-olds, or only one of the reports lists the drug *dronedarone*) comparing two distinct hypotheses regarding the two reports: i) they describe the same case; and ii) they describe cases that are independent of one another (23).

With a hit-miss model, matching information is rewarded with positive contributions that are greater the less frequent the matching values are (e.g. two reports matching on a rarely reported drug receive a larger score than two reports matching on a common drug like *paracetamol*).

Mismatching information is penalized, with greater penalties for features which seldomly mismatch on true duplicates (e.g., since there is greater coding ambiguity for adverse events than drugs, these are more likely to mismatch, even on true duplicates, and thus mismatching adverse events receive a lower penalty). However, these penalties do not depend on specific values (i.e., mismatching on *paracetamol* carries the same penalty as mismatching on a rarer drug).

The hit-miss model was previously extended to a hit-miss mixture model for numerical features and with a compensation for correlations between binary features (8), which are here applied to patient age and adverse event onset date (hit-miss mixture models) and to drugs/adverse events (correlation compensation).

### 3.3.4 Hit-miss models with country-specific drug and adverse event rates

The hit-miss model for drugs and adverse events in vigiMatch2017 use the overall reporting frequencies of drugs and adverse events in VigiBase to determine the reward/penalty. However, this assumes that these reporting frequencies are globally consistent, however medicines in routine use in high income countries may not be readily available in lower- or middle-income countries for example.

For the hit-miss models related to drugs and adverse events, vigiMatch2025 instead considers the reporting frequencies in the country of origin. Where two reports are not from the same country, the global frequencies are used. Parameters of the hit-miss models derived from known duplicates were shared across all countries, due to insufficient training data to infer them per country.

### 3.3.5 Date embeddings

Reports often include multiple dates and a shortcoming of vigiMatch2017 is that it considers only a single date of onset per report and will not reward matches on multiple distinct dates. Of particular interest for duplicate detection are the start and end dates of reported adverse events and drug therapies. Some of these are reported in structured fields, and others are described in the free text narrative of reports. We therefore developed a set of regular expressions to extract and normalize dates from the free text narrative.

In VigiBase, dates are stored as pairs of timestamps reflecting their uncertainty interval (e.g., a report of "March 2021" would be represented by 2021-03-01 00:00:00 and 2021-03-31 23:59:59)[1]. For our date embedding feature, we included dates which had a maximum uncertainty of 7 days and excluded all dates reported as 1st January, regardless of their uncertainty, since this was historically used by some organizations to indicate missing information on the month and day.

We embedded the dates by producing a one-hot-encoded vector for a report, such that each element of the vector represented a single day between 1900-01-01 and 2050-12-31, with the element being 1 if a report includes a date interval starting that date and 0 otherwise. All dates mentioned on a report were included in the same vector. We applied a convolution with a window size of 7 days, to partially reward near matches.

[1] A similar principle is applied in the treatment of patient age in its hit-miss mixture model

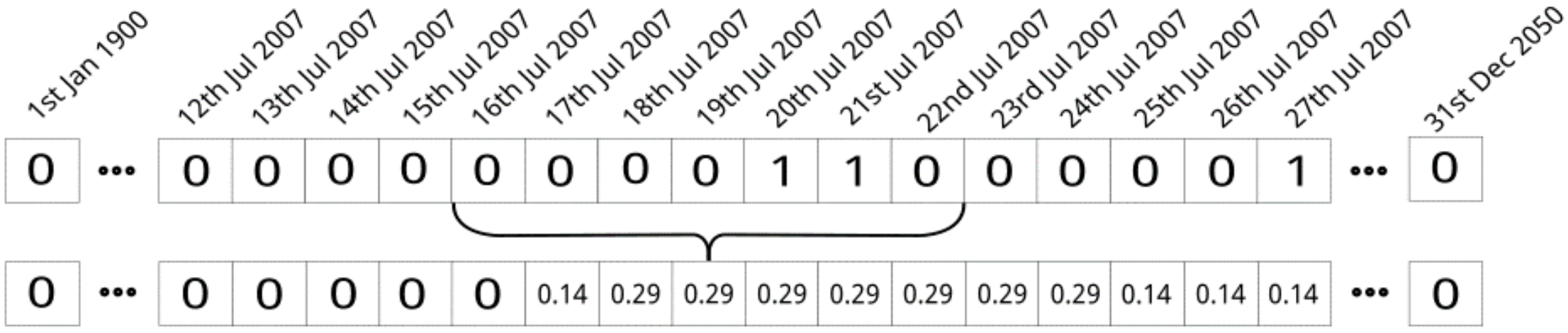

*Figure 1: Schematic of how the date vector would be constructed for a report mentioning 20th, 21st and 27th July 2007. The top row shows the one hot encoding, and the bottom row vector after applying a convolution with a window size of 7 days.*

The date vectors of two reports were compared using a cosine similarity score.

### 3.3.6 Externally indicated reports

As part of the E2B(R2) format, cases can be linked to previous transmissions via field A.1.11.2 (20). If a report had an ID in field A.1.11.2 (Case Identifier in the Previous Transmission) which exactly matched with another report's value in A.1.0.1 (Sender's safety report unique identifier), A1.10.1 (Regulatory authority's case report number) or A.1.10.2 (Other sender's case report number), we defined that pair as "externally indicated". We encoded this as a binary feature where the value is 1 if a pair is externally indicated, and 0 otherwise. We encoded this information as a feature, rather than use it as a heuristic, since a preliminary investigation estimated its precision in linking reports to be 0.76.

To avoid ID collisions due to senders using similar ID schemes, we limited matches to IDs starting with one of 30 strings clearly delineating a country (e.g., 'DE-', 'GB-', 'FR-', 'IT-' etc.)[2]. With this restriction, approximately 65,000 pairs were linked in this way.

### 3.3.7 Benchmark comparator method

As the benchmark comparator method for this study, we use vigiMatch2017. It largely follows the description in Norén et al (2007) (8), using the hit-miss model and its extensions to compute contributions to an overall match score from each of the following report elements: *patient age*, *patient sex*, *reporting country*, *date of onset* (for the adverse event), *outcome,* all *drugs* listed on the report (at WHODrug Active Substance level), all *adverse events* listed on the report (at MedDRA Preferred Term level), and *patient initials*[3].

Whereas vigiMatch2025 uses linear SVMs to obtain the overall match score, vigiMatch2017 assumes conditional independence between its features and obtains its total match score through summation of hit-miss model weights (which correspond to log-likelihood ratios). An explicit compensation for correlations between reported adverse events and drugs is included as described above but based on global reporting rates.

The original publication assumed a mixture of normal distributions for the match scores of duplicates and non-duplicates, respectively, to compute a threshold for suspected duplicates (8). Over time, this resulted in a drift toward ever higher thresholds, and vigiMatch2017 therefore, heuristically, uses the mean match score for known duplicate pairs as its threshold for suspected duplicates. A separate heuristic in vigiMatch2017 not included in the original publication requires that the net contribution

---

[2] A full list can be found in the supplementary material (S2).

[3] vigiMatch2017 is integrated in the processing of incoming data and can leverage some information not included in VigiBase

from patient age, patient sex, date of onset, and patient initials be greater than 0, for two reports to be flagged as suspected duplicates.

Due to high observed false positive rates when applying vigiMatch2017 to vaccine reports during its routine application, it has subsequently only been applied to drug pairs and so can only act as a comparator to the vigiMatch2025 drug model.

## 3.4 Performance evaluation

### 3.4.1 Precision Studies

Precision is a measure of how reliable a model's predictions are. It is defined as.

$$Precision = \frac{True\ Positives}{True\ Positives\ +\ False\ Positives}$$

The *FDAGold*, *FDASilver*, *AEMPS*, *Lareb* and *MHRA* reference sets contain no non-duplicates and so they cannot be used to estimate the model's precision. The UMC reference set contains some labelled non-duplicates, but the biases in how the dataset was built make it unsuitable for measuring precision.

To estimate the precision of the models, a stream of random pairs was presented to each model until it had classified 100 pairs as suspected duplicates. Random pairs were generated from the same sequence of random seeds for each model, in batches of $10^8$ pairs. These predicted duplicates were examined and annotated by a medical doctor (author JFC). In this process, report pairs were labelled as either duplicates, otherwise related or as non-duplicates according to the annotation guideline (supplementary material S1). We present precision results both for duplicates, and for related pairs (being either duplicates or otherwise related). Author JWB also labelled all pairs to measure inter-annotator agreement, however JFC's annotations were considered authoritative.

These labels were additionally used to estimate the expected number of true duplicates detected per report in the dataset. This was computed as

$$Expected = \frac{Number\ of\ True\ Duplicates\ Found}{Number\ of\ Random\ Pairs\ Compared} \cdot (Number\ of\ Reports\ in\ Dataset - 1)$$

Where the number of reports in the dataset minus 1 is the number of pairs per report. This measure gives an indication of the prevalence of detectable duplicate pairs for each model. Moreover, since vigiMatch2017 and vigiMatch2025 were applied to the same stream of random pairs, a model with a higher rate of detecting true duplicates can be interpreted as having a higher recall.

We also evaluated the performance of the models on report pairs from 3 individual countries which each had few enough reports that we could exhaustively run on all possible pairs within a reasonable time. 2 of these were African countries which are known to have adverse event and drug distributions which are different from the global pattern. We also chose 1 European country. 15 random pairs predicted as duplicates were selected for each country, for each model, and a medical doctor (author JFC) annotated them according to our guideline. We report precision for each country and model. We also computed the number of reports that would remain if only one representative report was kept per *duplicate group.* A duplicate group is here defined as a set of duplicate pairs that are fully connected, so that every report in the group is considered a duplicate of every other report in the group. We find these groups using the networkx Python library version 3.2.1.(24)

### 3.4.2 Recall on Labelled Datasets

We estimated the models' recall by applying them to the *test* labelled datasets described in table 1, and computing recall as

$$recall = \frac{True\ Positives}{True\ Positives + False\ Negatives}$$

Since some of the labelled datasets were partially built using E2B(R2) field A.1.11.2, we also report recall after artificially setting the externally indicated feature value to 0, i.e., as if no pairs were externally indicated.

# 4 Results

## 4.1 Trained Models

After training the drug and vaccine models, we investigated the feature importance. Since features were on different scales, the $\vec{w}$ coefficients were not meaningful independently. We thus looked at each feature's contribution to $\vec{w} \cdot \vec{x}_i$ for known duplicates among the training and validation pairs, and for 1,000,000 random pairs.

In figure 2 we display box plots representing the distribution of contributions from each feature for each model. Pairs excluded by a heuristic are not shown. The distribution for known duplicates is displayed in green, and for the random pairs in purple. For both models, the true duplicates tend to receive higher scores compared to random pairs. The contribution from the drug/AE hit-miss model is typically the largest. Demographic features (i.e., age, sex) contribute less to the scores for vaccines than they do for drugs, which is in line with intuition since vaccine recipients tend to be more demographically homogeneous. Mismatches on sex are also much more heavily punished for the vaccine model.

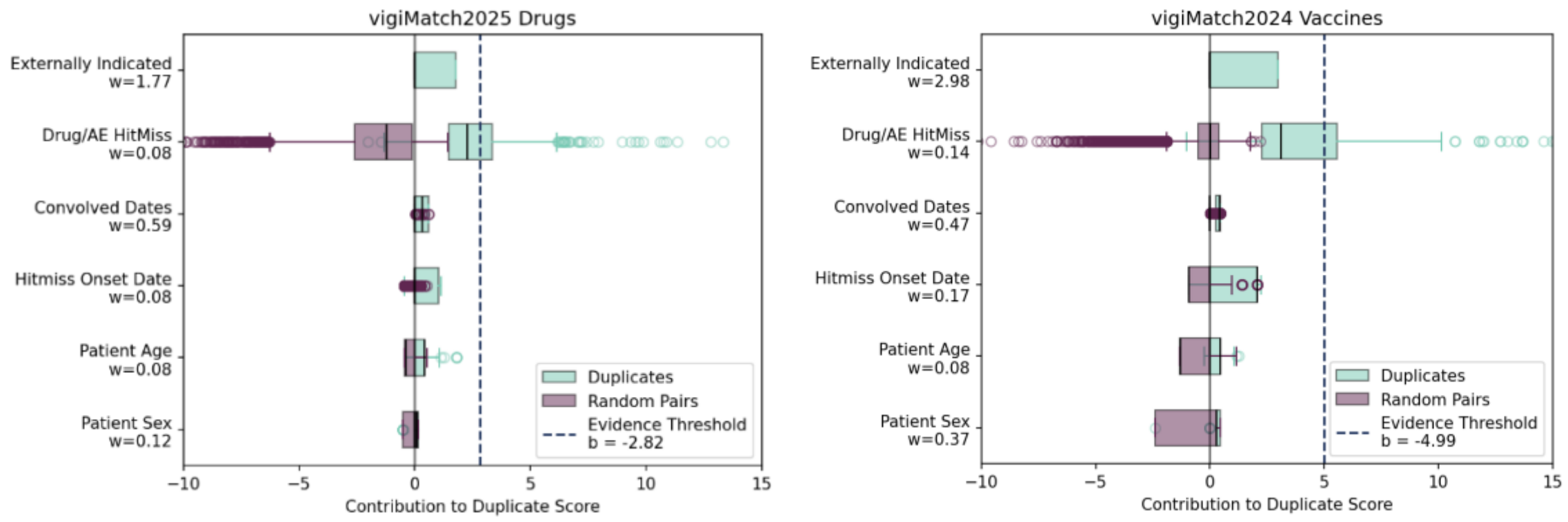

*Figure 2: The feature weights and contributions to $\vec{w} \cdot \vec{x}_i$ for true duplicates and random pairs. The fitted SVM weights are given with the y tick labels. Note some extreme outliers for the Drug/AE HitMiss feature are outside the range of these plots.*

## 4.2 Precision Studies

The results for each model are presented in Figure 3 and Table 2. vigiMatch2025 Drugs achieves higher precision than vigiMatch2017, and vigiMatch2025 Vaccines achieves a higher precision still. As seen in Figure 4, the expected number of true duplicates per report is higher for vigiMatch2025 drugs than vigiMatch2017, which indicates a higher recall. vigiMatch2025 Vaccines has a comparable, but slightly lower expected number of true duplicates per report. The inter-annotator agreement was a Cohen-Kappa score of 0.67, indicating moderate to substantial agreement (25).

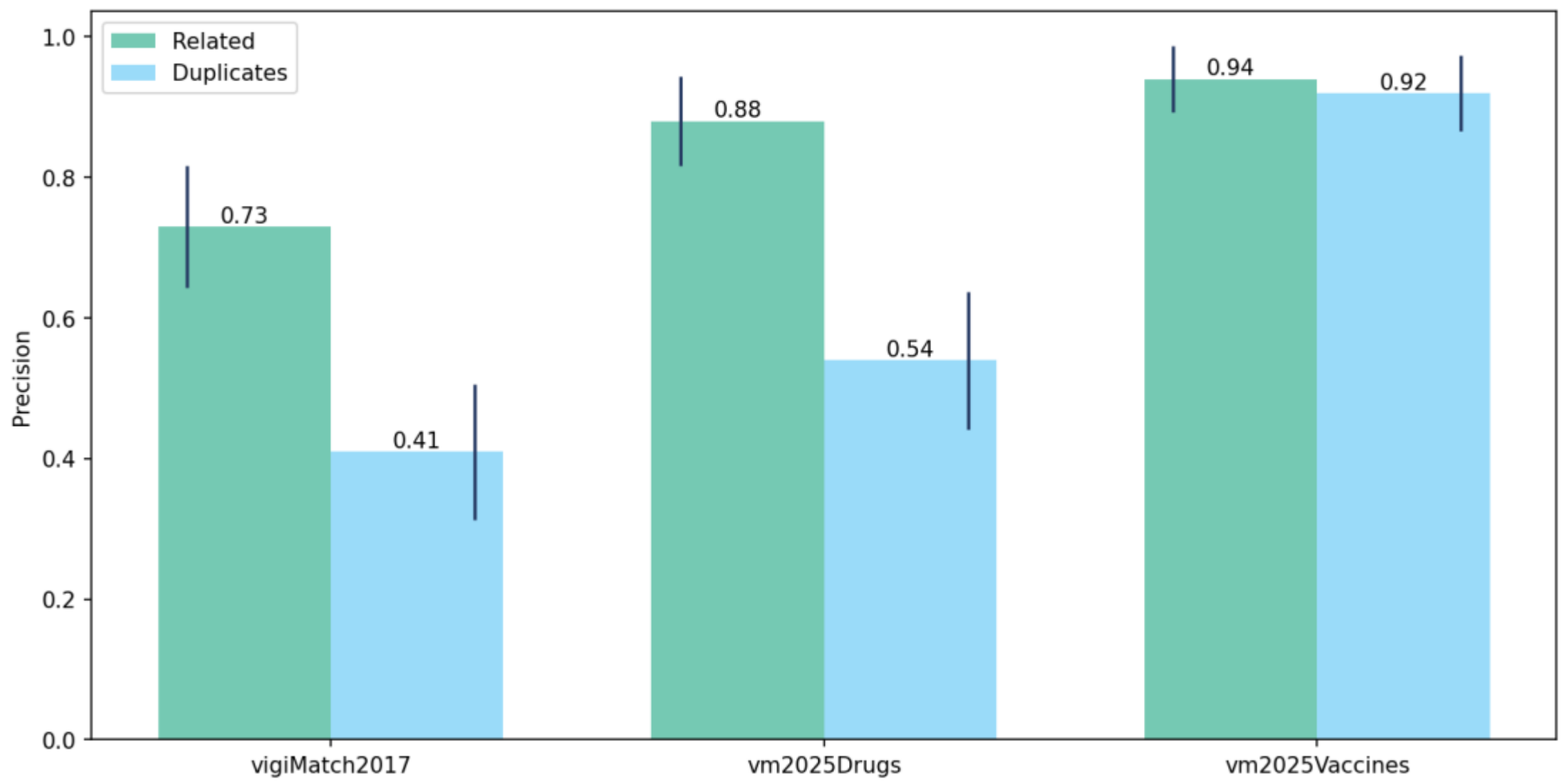

*Figure 3: Results from the evaluation of the models' precision. Vertical lines represent the 95% confidence interval estimated with Wald's method. The blue bars represent the true duplicates the model is trained to classify. The green bars are reports which are either duplicates, or otherwise related.*

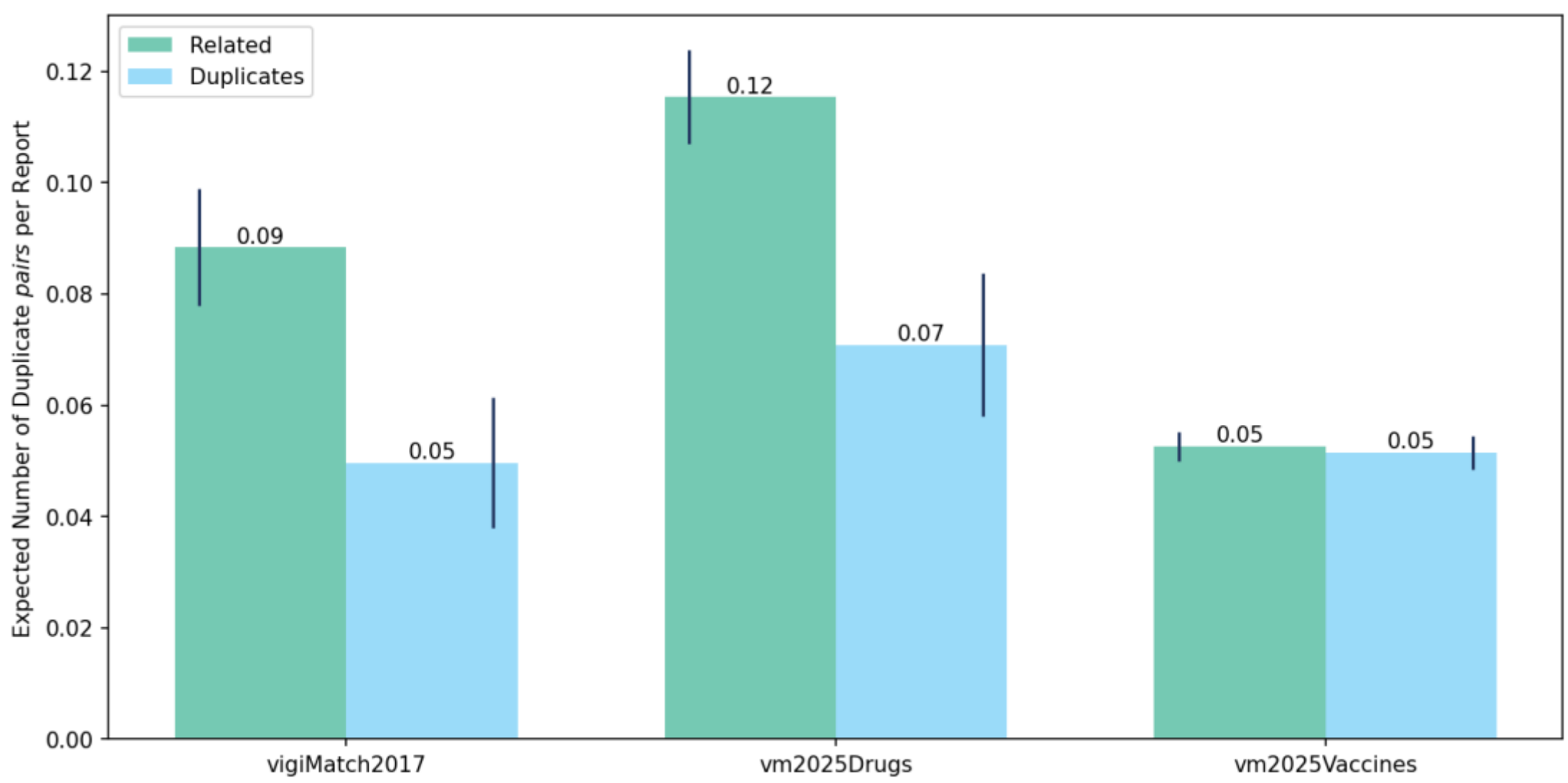

*Figure 4: The expected number of true duplicates found per report for each model, with vertical lines representing the 95% confidence interval estimated with Wald's method. The blue bars represent the true duplicates the model is trained to classify. The green bars are reports which are either duplicates, or otherwise related.*

Of the 100 pairs predicted as duplicates, one (for drugs) and four (for vaccines) had positive contributions from the externally indicated feature. All these five pairs were true positives and remained as true positives when masking the contribution from the externally indicated feature.

| Model | N Reports (Millions) | N Random Pairs Compared (Billions) | N Predicted Duplicates | N True Positives | Precision | True Positives per Billion Pairs | True duplicates detected per report |
|---|---|---|---|---|---|---|---|
| VigiMatch2017 | 26.9 | 22.2 | 100 | 41 | 0.41 | 1.85 | 0.05 |
| VigiMatch2025 Drugs | 26.9 | 20.5 | 100 | 54 | 0.54 | 2.63 | 0.07 |
| VigiMatch2025 Vaccines | 1.8 | 3.2 | 100 | 92 | 0.92 | 28.75 | 0.05 |

*Table 2: Tabulated results from the precision experiments.*

The results for the individual countries are in table 3. For the two African countries, vigiMatch2017 predicts significantly more pairs as duplicates than vigiMatch2025, however there were no true positives in the two times 15 pairs sampled, whereas vigiMatch2025's precision was commensurate with the results in figure 3. The results for vigiMatch2017 and vigiMatch2025 were roughly equal for the European country, with vigiMatch2017 predicting slightly more pairs as duplicates, and finding two additional true positives. It's also seen that vigiMatch2025 removes significantly fewer reports via deduplication than vigiMatch2017 for both African countries.

| Country | African Country A | African Country B | European Country A |
|---|---|---|---|
| **N Reports in VigiBase** | 19,042 | 4,363 | 1,004 |
| **vigiMatch2017 Remaining Reports after removing suspected duplicates** | 13,363 | 3,503 | 976 |
| **vigiMatch2025 Remaining Reports after removing suspected duplicates** | 18,756 | 4,243 | 987 |
| **N Possible Pairs (Millions)** | 181.3 | 9.5 | 0.5 |
| **vigiMatch2017 Predicted Duplicate Pairs** | 42,933 | 1,969 | 30 |
| **vigiMatch2025 Predicted Duplicate Pairs** | 409 | 137 | 21 |
| **vigiMatch2017 precision** | 0/15 | 0/15 | 8/15 |
| **vigiMatch2025 precision** | 6/15 | 13/15 | 6/15 |

*Table 3: Results of the individual country experiments, showing each model's precision on pairs classified as duplicates after exhaustively classifying all pairs in VigiBase from that country.*

## 4.3 Recall

Recall for each of the *test* reference sets is presented in Figure 5. vigiMatch2025 Drugs achieves a higher recall for every dataset, both with and without relying on the externally linked feature. The recall for the vigiMatch2025 Vaccines is comparable to the vigiMatch2025 drug model for most datasets.

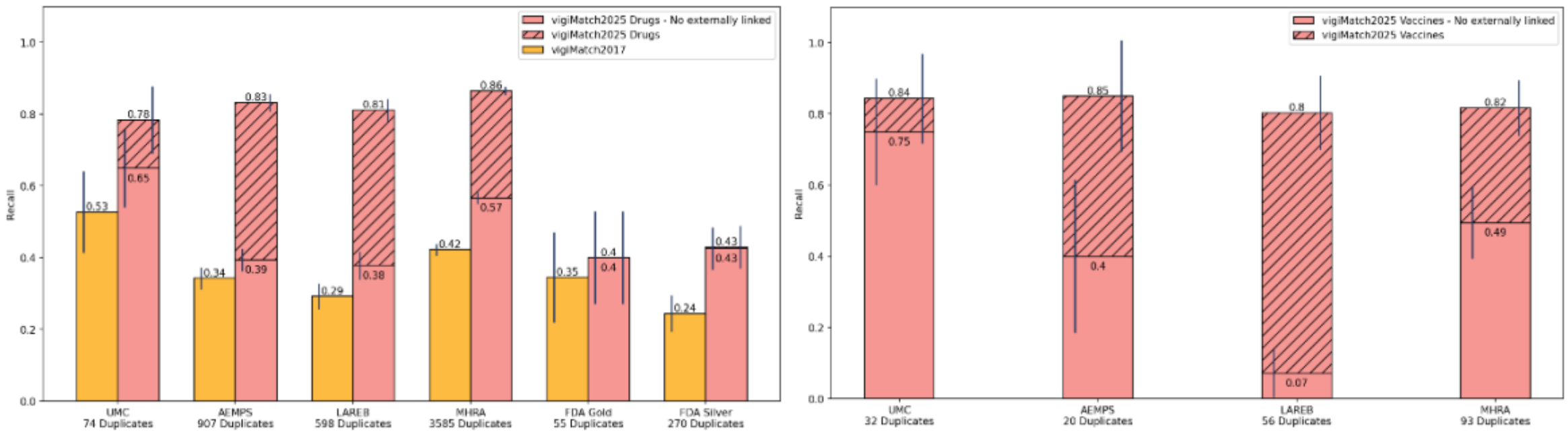

*Figure 5: Recall performance of each of vigiMatch2017, vigiMatch2025 Drugs and vigiMatch2025 Vaccines. The hatched bars represent the complete performance of the model, and the solid bars represent the performance by assuming that no pairs can be linked via the reported duplicate IDs. 95% confidence intervals for the height of each bar are presented in dark blue, computed using Wald's method.*

## 4.4 Qualitative Review of Cases

A qualitative review was conducted to give a better understanding where vigiMatch2025 improves upon vigiMatch2017 and where it makes mistakes. A selection of true positives, false positives, false negatives and true negatives from the test reference set were examined. The examples where vigiMatch2025 succeeded over vigiMatch2017 reflected known limitations of vigiMatch2017, like its inability to account for multiple dates on each report and country-specific reporting patterns. The cases where vigiMatch2025 fails tended to be unusual cases, such as non-duplicates with many matching adverse event terms. The qualitative review is described in detail in supplementary material S3.

# 5 Discussion

vigiMatch2025 demonstrates a substantial improvement in performance over the previous state of the art comparator model vigiMatch2017, whilst retaining many attractive features of the earlier model. Importantly, the new model achieves improved precision and recall for duplicate detection of both drug and vaccine reports in VigiBase, with a much lower false positive rate for certain countries.

The training set for vigiMatch2025 has been expanded with over 1500 additional labelled pairs. This supports its incorporation of separate SVM predictive models for drugs and vaccines, which allows it to adapt its weights based on the most relevant examples in training data. This contrasts with vigiMatch2017 which leans heavily on the assumed stochastic model and overall relative frequencies of different reporting elements in the database. However, this increased adaptability comes with an increased risk of overfitting, if the training data are selected improperly or later become unrepresentative of duplicates in VigiBase. Another advantage of vigiMatch2025 is that its hit-miss models for drugs and adverse events are trained per country. We saw a significant reduction in the false positive rates in duplicate detection for countries with drug and adverse event distributions that differ from those in the global database. vigiMatch2025 also considers all unique dates related to drugs and adverse events on a report including in the case narrative. This holistic approach to dates allows for a more nuanced comparison of the timelines presented in each report.

At the same time, vigiMatch2025 retains the explainability of vigiMatch2017. The choice of a linear kernel for the SVM makes the prediction a simple weighted sum of the features and constant intercept, which can be interpreted as an accumulation of evidence and an evidence threshold. The model only classifies a pair as duplicates when there is sufficient evidence. A specific prediction by the model can also be easily interrogated for which elements of the report pair contributed to the model's prediction. The evidence threshold can also be adjusted to favor precision or recall, depending on the application.

vigiMatch2025 retains the low computational cost of vigiMatch2017, which is crucial to most applications. The new features all scale equally well with the number of reports in a database as vigiMatch2017 does, which in the production environment currently compares around 70 million pairs of reports each second. The new representation of dates only requires a dot product of two very sparse vectors at inference time. The country specific drug and adverse event hit-miss models do not represent any change in computational complexity at comparison time over the global model, as they just use different parameters for the same calculations.

The acquisition of unbiased training data for duplicate detection models poses a significant challenge, due to the extremely low prevalence of positive examples. The annotations themselves also presented a challenge. VigiBase is a highly diverse dataset, and the types of evidence of duplication observed during annotation was equally diverse. Our annotation guideline reflects this diversity of evidence, with room left for expert judgement when reaching a final decision for an annotation. The training data used in this study was accumulated during development of vigiMatch2025, meaning that biases undoubtedly exist. The evaluations presented in this paper were constructed to mitigate them as much as possible. The precision experiments, by design, give unbiased estimates of the models' performance. Several different, independent datasets, from the FDA, AEMPS, MHRA, and Lareb national pharmacovigilance centres were used to evaluate the models' recall. However, in future studies, an even more diverse collection of test sets may further probe models' recall. Moreover, the provenance of the pairs in these datasets, and on what basis they were annotated as duplicates, is not always known.

National centres often have access to much more information on each report than is included in the anonymized versions shared in VigiBase. Some of the pairs of true duplicates provided for the recall studies, do not have sufficient information in VigiBase to be annotated as duplicates under our guideline. In other words, even a human annotator with access only to VigiBase cannot achieve perfect recall against these reference sets. It also means that comparison to duplicate detection algorithms that leverage the richer information available in a national database is difficult, even when evaluating against the same reference set (e.g., the FDA silver and gold datasets are also used in (10)). The performance of vigiMatch2025 should be comparable on smaller or more localized databases. Indeed, in such settings, including the additional available information as features to the model would likely improve performance. This could also facilitate duplicate detection at time of data entry or report submission, which begs further development and testing in future studies.

The externally indicated feature is a strong predictor in vigiMatch2025, and by construction it was highly prevalent in the test sets from Lareb, AEMPS, and MHRA, whilst the precision evaluations highlighted that such a link is relatively rare in general. Indeed, there are only 65,000 pairs in VigiBase that can be linked in this way. However, due to the substantial improvement in performance where the feature was applicable and the absence of significant performance reduction elsewhere, we decided to retain it.

Areas for future research might include investigation of more advanced blocking schemes, to further decrease the computational burden. Additional features such as drug dosages may also be explored. Their low prevalence makes them difficult to incorporate effectively in a generalist model, however the human annotators acknowledged their value as evidence for or against duplication. There is additionally some scope to improve the representation of report dates in the model. Whilst the date embedding vectors successfully capture a more comprehensive and nuanced picture of a report's timeline, they do not penalize mismatches more than missing dates, nor account for the number of dates matching on a report. The hit-miss model for onset date was retained in the model to partially mitigate this shortcoming, however it would be desirable in future models to capture date information within one feature.

In our annotation, we used more granular labels, also noting whether the reports in a pair were likely otherwise related. That many of the false positives in the precision evaluation were otherwise related suggests that the mistakes the model makes are generally understandable. However, this also raises the possibility of further refining the model with a second classifier to distinguish between otherwise related and duplicate reports. Such a model would not be constrained to be so computationally inexpensive, and so more sophisticated models, such as large language models, could be considered.

# 6 Conclusion

Our study introduces a new predictive model, vigiMatch2025, for duplicate detection in databases of adverse event reports. We demonstrate that the model outperforms the previous state of the art model, vigiMatch2017, in every way that we measured, whilst retaining desirable properties such as low computational cost and explainability.

# Declarations

## Acknowledgements

The authors are indebted to the members of the WHO Programme for International Drug Monitoring who contribute reports to VigiBase. However, the opinions and conclusions of this study are not necessarily those of the various member organisations nor of the WHO. MedDRA® trademark is registered by ICH. Nils Erlanson, as of September 2023, is no longer employed by UMC. However, his contributions to the study were made prior to the time of departure, as part of their employment at UMC.

## Data Availability

The data that support the findings of this study are not publicly available. Access to the data is restricted based on the conditions for access within the WHO Programme for International Drug Monitoring. Subject to these conditions, data is available from the authors on reasonable request.

## Ethics approval and consent to participate

Not applicable. This study did not use personal data or human subjects.

## Competing interests

The authors declare that they have no competing interests.

## Declaration of generative AI and AI-assisted technologies in the manuscript preparation process

During the preparation of this work the author(s) used Claude Opus 4.5 and ChatGPT 5.2 in order to review sections of the manuscript and propose refinements, as well as perform a holistic review of the article for consistency. After using this tool/service, the author(s) reviewed and edited the content as needed and take(s) full responsibility for the content of the published article.

## S1: Annotation Guideline

**General principles**

Based on all information available in VigiBase for the two reports make your best judgment, as if you had to place a bet, on one of the three categories. In most cases, it will not be possible to know for sure (without going back to the national centre for additional information, which is out of scope here) if two reports are duplicates or non-duplicates.

**Considerations in assessing duplicates**

- Specific information in narratives can be very valuable both in providing evidence for and against two reports being duplicates, though beware that otherwise related reports can have near-identical narratives as well
- Independent reports are unlikely to match on exact dates – especially so, when there are multiple matching dates linked to the same event – for example an adverse event onset or a drug start
- Matches on multiple drugs and adverse events can be unlikely for independent reports, especially if they are not a common combination of drugs and/or adverse events
- For decision support, you could consider how common a specific combination of drugs and/or adverse events is in VigiBase via Vigilyze's search filter (remember the 'and' connection)
- If there is matching information of a general nature such as the drugs and adverse events or specific numerical features from the narrative but demographic information such as patient age and sex differs, reports may be more likely to be otherwise related than duplicates (e.g. same reporter, associated with a litigation case, reports of separate adverse events in the same patient, ...)

**Categories**

- Non-duplicates
  - Pairs of reports that should not be linked at all
- Flag - possible duplicates
  - Reports that you would like to flag to an assessor as possible/likely duplicates for their consideration in reviewing a case series

- Reports that you would be fine with feeding back to a member organisation as suspected duplicates
- Reports that you would like to have the option to exclude in statistical signal detection

- Flag - Otherwise related reports
  - Reports that are not independent of one another, but also not duplicates
  - Reports for the same patient but different (occurrences of) adverse events
  - Reports from the same primary reporter
  - Reports for members of the same family
  - Reports from the same study / article
  - Reports related to the same litigation
  - Reports associated with e.g. the same product quality, contamination, cold chain issue etc
  - Reports of adverse event happening at the same mass administration event (e.g. the vaccination of the same class of school children),
  - … but <u>not</u> reports just generally associated with the same mass administration campaign

## S2: Externally Linked Country Strings

When linking pairs via E2Br2 field A.1.11.2, only identifiers starting with one of the following strings were considered

'DE-', 'GB-', 'FR-', 'IT-', 'NL-', 'GR-', 'BE-', 'PT-', 'CA-', 'ES-', 'PL-', 'CZ-', 'JP-', 'US-', 'DK-', 'FI-', 'AT-', 'AU-', 'SE-', 'PHH', 'RO-', 'IE-', 'CH-', 'HU-', 'NO-', 'IN-', 'SK-', 'BR-', 'HR-', 'EG-'

## S3: Illustrative Examples of Performance

### Drugs - True Positive

The following example is of a pair of reports which are a true duplicate pair identified by vigiMatch2025 but not by vigiMatch2017. The reports both describe the outcome of a patient involved in an study on platelet aggregation inhibition by Eptifibatide or Tirofiban during primary percutaneous interventions (1). The reports contain relatively little structured information about the patient. However, it is clear from the narrative that the reports are describing the same fatal case from the study. Crucially, vigiMatch2025 is able to make use of the dates described in the narrative, as well as the fact that the reports were linked in the e2b field A.1.11.2 as being duplicates, both of which vigiMatch2017 does not have access to, to add to the evidence from the matching drug and reaction and correctly identify the pair as a duplicates.

| | Report 1 | Report 2 | vigiMatch2025 Score |
|---|---|---|---|
| Sex | Not reported | Not reported | 0 |
| Age | Not reported | Not reported | 0 |
| Dates | 2 matching and one mismatching date in the narrative<br>No reaction dates reported | | 0.59<br>0 |
| Drugs/Reactions | 1 matching reaction<br>1 matching drug | | 2.29 |
| Externally Linked | Yes | | 1.76 |
| Total Score | | | 4.63 |
| Patient Heuristic | PASS | | 0.59 |

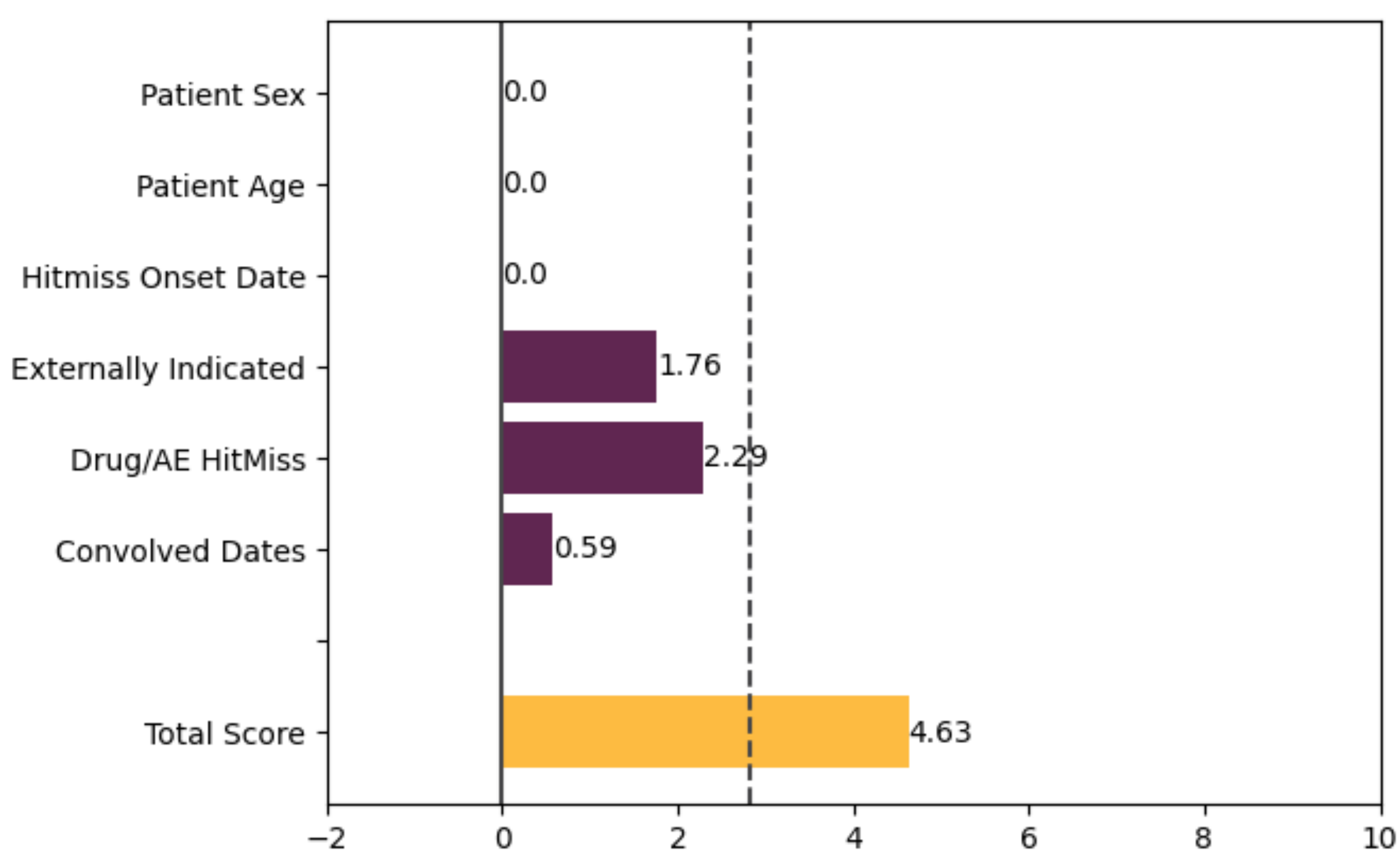

## Drugs - False Negative

The following example is a of a pair of reports which are a true duplicate pair identified by vigiMatch 2017 but not by vigiMatch2025. The reports both come from a high population country in Europe with many reports in VigiBase. The age appears to have been miscoded in one of the reports, since the patient is described in both free text narratives as being 48, leading to a significant negative contribution to the vigiMatch score. This leads to the pair failing to pass the patient heuristic, and thus being classified as non-duplicates, The base model would have otherwise classified them as a duplicate. It is possible that one describes the age at onset of the first reaction, and the other the age at the time of reporting, since the narratives describe a clinical course spanning several years. Moreover, whilst both narratives describe, in detail, the same clinical course, since one report seems to be a follow-up of the other, there are several dates, mostly relating to laboratory tests, in one narrative which don't appear in the other, leading to a relatively low score from the dates, despite many matches in the reaction and drug fields.

| | Report 1 | Report 2 | vigiMatch2024 Score |
|---|---|---|---|
| Sex | Male | Male | 0.15 |
| Age | 42-43 years | 48-49 years | -0.46 |
| Dates | | | 0.2<br>1.04 |
| Drugs/Reactions | 3 matching reactions<br>1 mismatching but semantically similar reaction on each report<br>5 distinct matching drugs | | 4.22 |
| Externally Linked | Yes | | 1.76 |
| Total Score | | | 6.93 |
| Patient Heuristic | FAIL | | -0.1 |

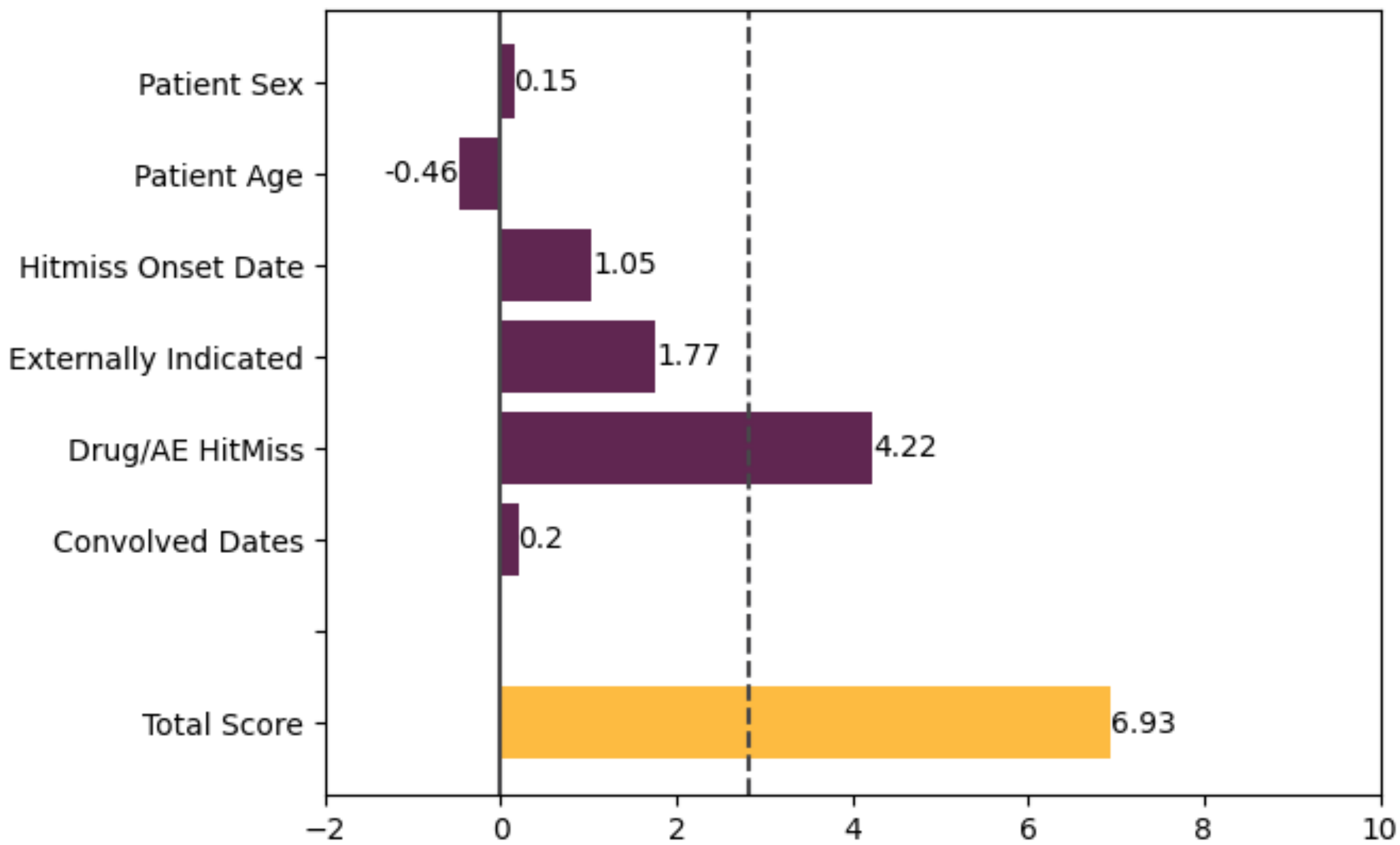

## Drugs - False Positive

The following example is a pair of reports which are not considered a duplicate pair under our annotation guideline, but which vigiMatch2025 identifies as a suspected duplicate. Both reports come from a large population country in the Americas with many reports in VigiBase. These cases are practically identical, and very well might be related, however do not contain enough evidence to label them as such according to our annotation guideline.

| | Report 1 | Report 2 | vigiMatch2025 Score |
|---|---|---|---|
| Sex | Male | Male | 0.15 |
| Age | Not reported | Not reported | 0 |
| Dates | No reported dates | | 0<br><br>0 |
| Drugs/Reactions | 2 matching reactions<br>3 matching drugs | | 2.78 |
| Externally Linked | No | | 0 |
| Total Score | | | 2.93 |
| Patient Heuristic | PASS | | 0.15 |

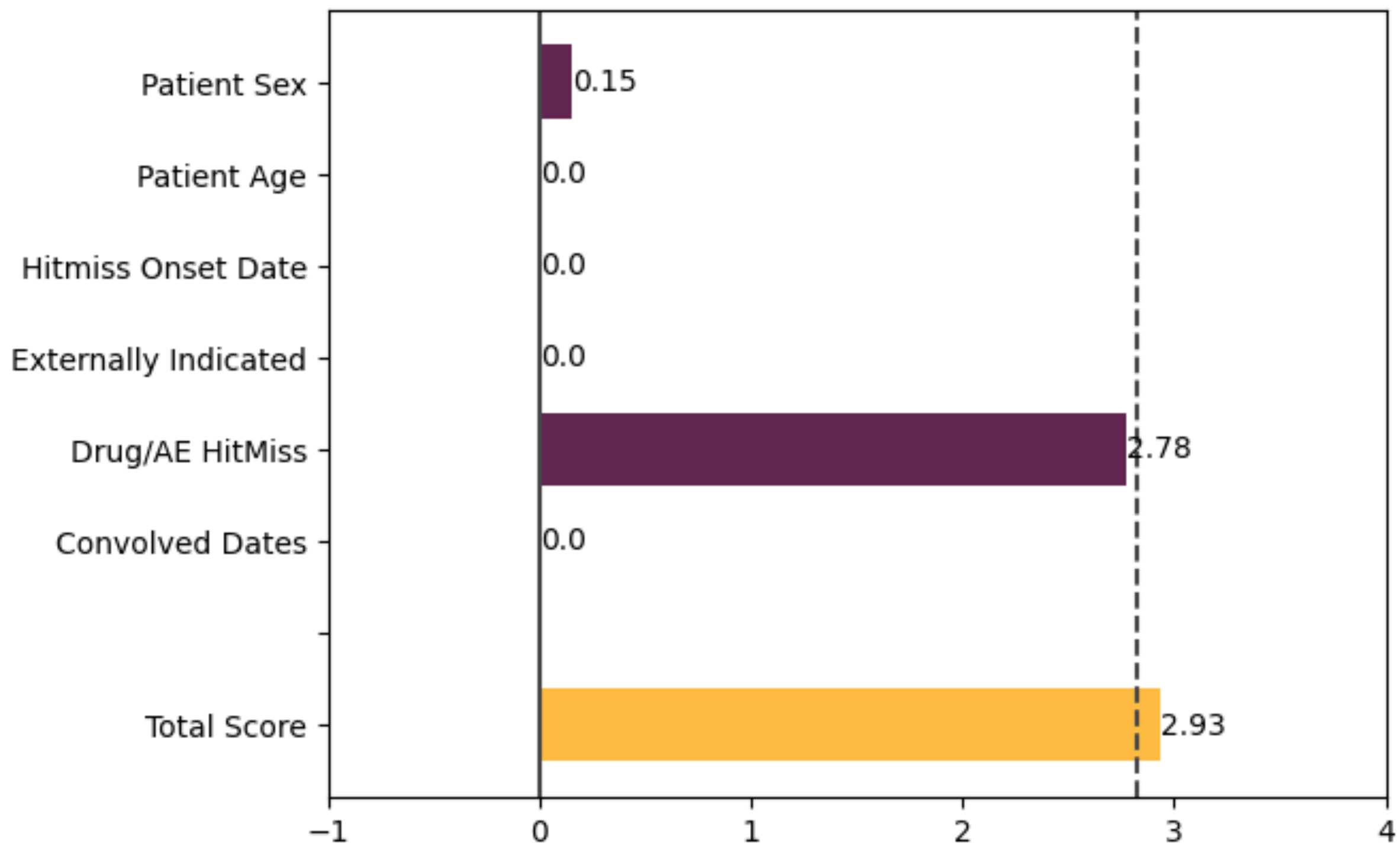

## Drugs - True Negative

This example is a pair of reports correctly identified as non-duplicates by vigiMatch2025 and spuriously identified as duplicates by vigiMatch2017. The reports come from a high population country in Africa with relatively few reports in VigiBase. The reports describe two patients who both experienced the same reaction within a day of one another, as a suspected result of taking the same three medications. Since the country has few reports, and all of the drugs and reactions are globally uncommon, vigiMatch2017 therefore spuriously identified these as duplicates. However, in the context of the country, the drugs and reactions are in fact rather common. The reaction occurs on around 1/6 cases from this country, and 1/4 cases have at least one of the drugs. Moreover, the patients have different reported sex. Whilst not used in either model, the patients are also reported to have substantially different body weights. This pair doesn't make it past the patient heuristic, although it's key to note that the pair wouldn't have been considered a duplicate even if the sexes had matched.

| | Report 1 | Report 2 | vigiMatch2025 Score |
|---|---|---|---|
| Sex | Male | Female | -0.53 |
| Age | Not reported | 23-24 years | 0 |
| Dates | 2 matching distinct dates<br>Earliest reaction date mismatch by 1 day | | 0.25 |
| | | | 0.25 |
| Drugs/Reactions | 2 matching reactions<br>3 matching drugs | | 0.36 |
| Externally Linked | No | | 0 |
| Total Score | | | 0.33 |
| Patient Heuristic | FAIL | | -0.28 |

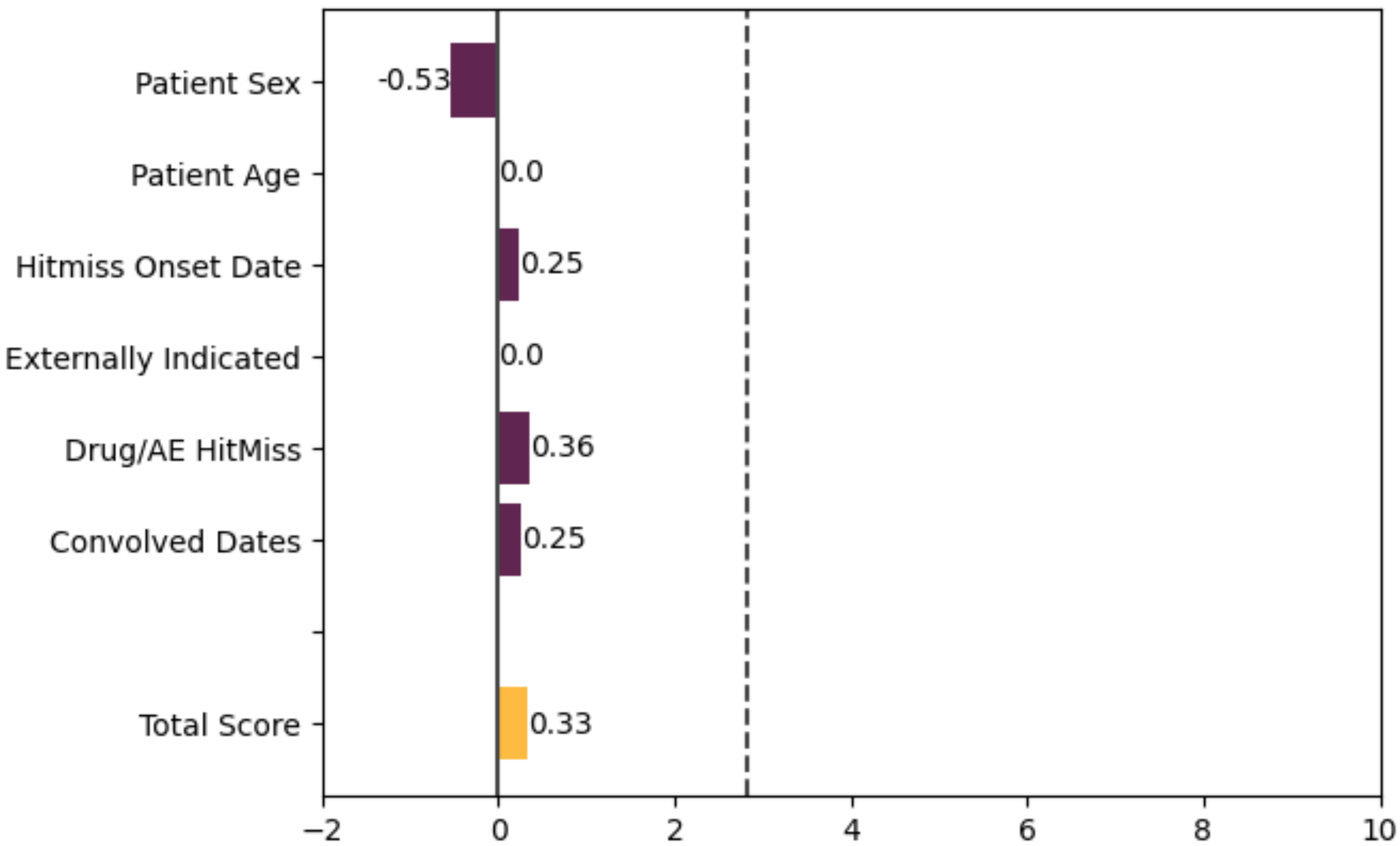

## Vaccines - False Positive

These two reports are identified as a suspected duplicates by vigiMatch2025 but are labelled as non-duplicates according to this paper's annotation guideline. The reports both come from a high population country in the Americas with many reports in VigiBase. These reports contain an unusually large number of reported reactions, the vast majority of which are representing diagnostic tests and their results (which are rare to be coded as reactions in VigiBase). The age hit-miss model also doesn't consider the magnitude of the patient age when determining the match score, where a mismatch of a few months should probably be much more significant in infants, like this case, than in adults. Here the ages are rewarded for being close.

| | Report 1 | Report 2 | vigiMatch2025Score |
|---|---|---|---|
| Sex | Female | Female | 0.29 |
| Age | 109-146 days | 36-73 days | 0.3 |
| Dates | 2 distinct dates on each report, none of which match<br>Earliest onset dates differ by 18 months | | 0<br>-0.93 |
| Drugs/Reactions | 29 matching reactions<br>86 mismatching reactions<br>4 matching vaccines | | 5.84 |
| Externally Linked | No | | 0 |
| Total Score | | | 3.14 |
| Patient Heuristic | PASS | | 0.59 |

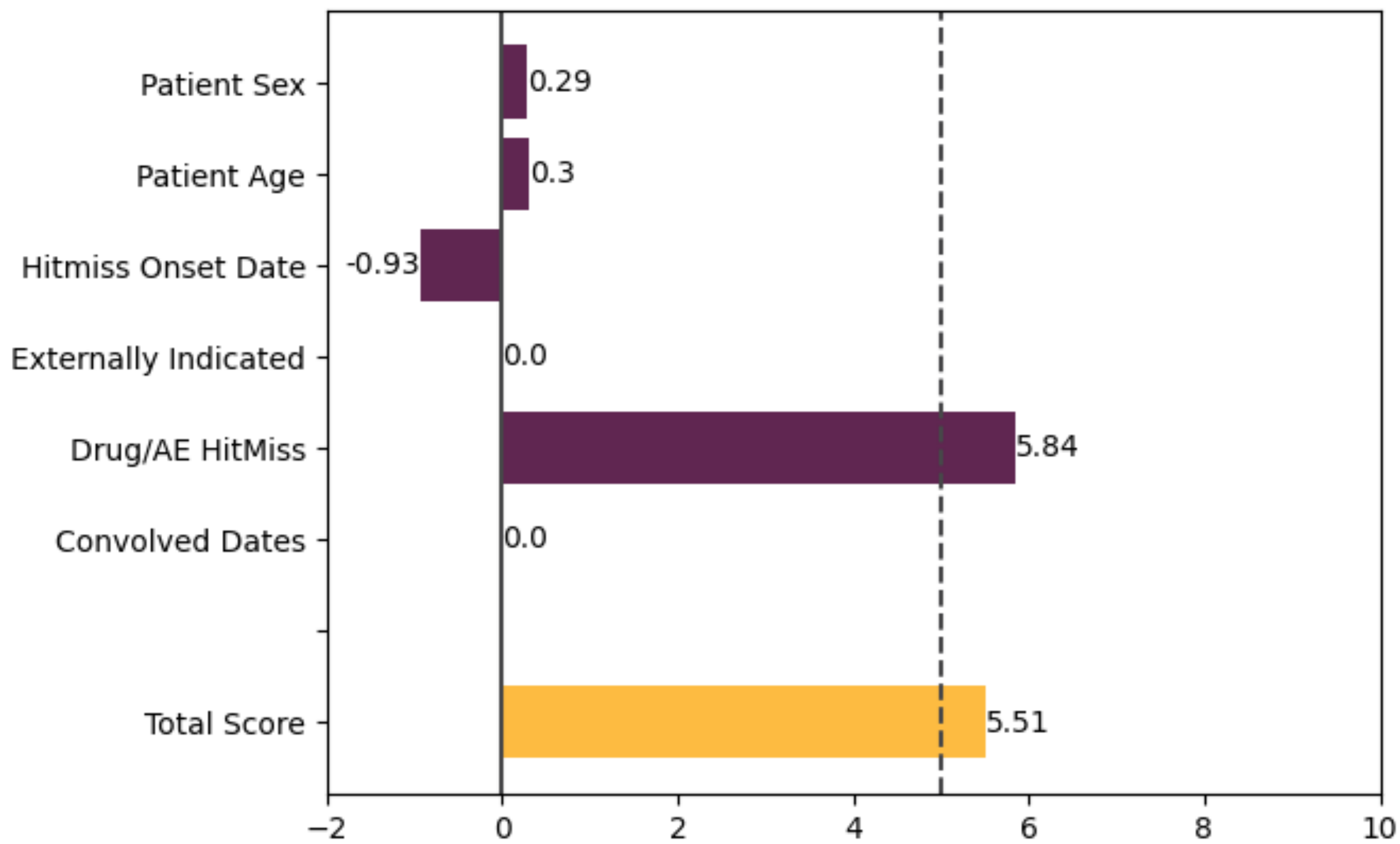

## Vaccines - False Negative

These two reports are labelled as a suspected duplicate pair by the human annotator, but vigiMatch2025 fails to identify them as such. The reports come from a medium sized European country with many reports in VigiBase. The pair appears to be a report from the first dose of a childhood schedule of vaccines, and the other report referring to the same adverse event but including information up until the third dose, so a possible follow-up. This pair exemplifies the high standard of evidence necessary for a vaccine pair to be suspected as a duplicate according to our annotation guideline, but it may also reflect a computational deviation. Whilst many of the drugs and reactions are matching, they have a high degree of correlation for this country. Before it is moderated by the SVM coefficient, the drug/reaction binary hit-miss model for this pair comprises:

- + 7.36 for the drug model
- + 9.95 for the reaction model
- - 22.19 for the correlation compensation

Leading to an overall negative contribution to the score. This behaviour is incorrect (the resulting score for a set of matching drugs and adverse events should always be positive), and arises from an edge case where the correlations between drugs and adverse events are extremely high and the drugs/adverse reactions make up a significant proportion of that country's dataset, leading to a breakdown in one of the approximations used in the computation of the correlation compensation.(2)

| | Report 1 | Report 2 | vigiMatch2025 Score |
|---|---|---|---|
| Sex | Female | Female | 0.29 |
| Age | Not reported | 1-2 years | 0 |
| Dates | 3 matching dates out of 5 total distinct dates<br>Earliest onset date matches | | 0.39<br>1.26 |
| Drugs/Reactions | 8 matching reactions<br>3 mismatching reactions<br>3 matching vaccines/drugs | | -0.67 |
| Externally Linked | Yes | | 2.98 |
| Total Score | | | 4.26 |
| Patient Heuristic | PASS | | 0.68 |

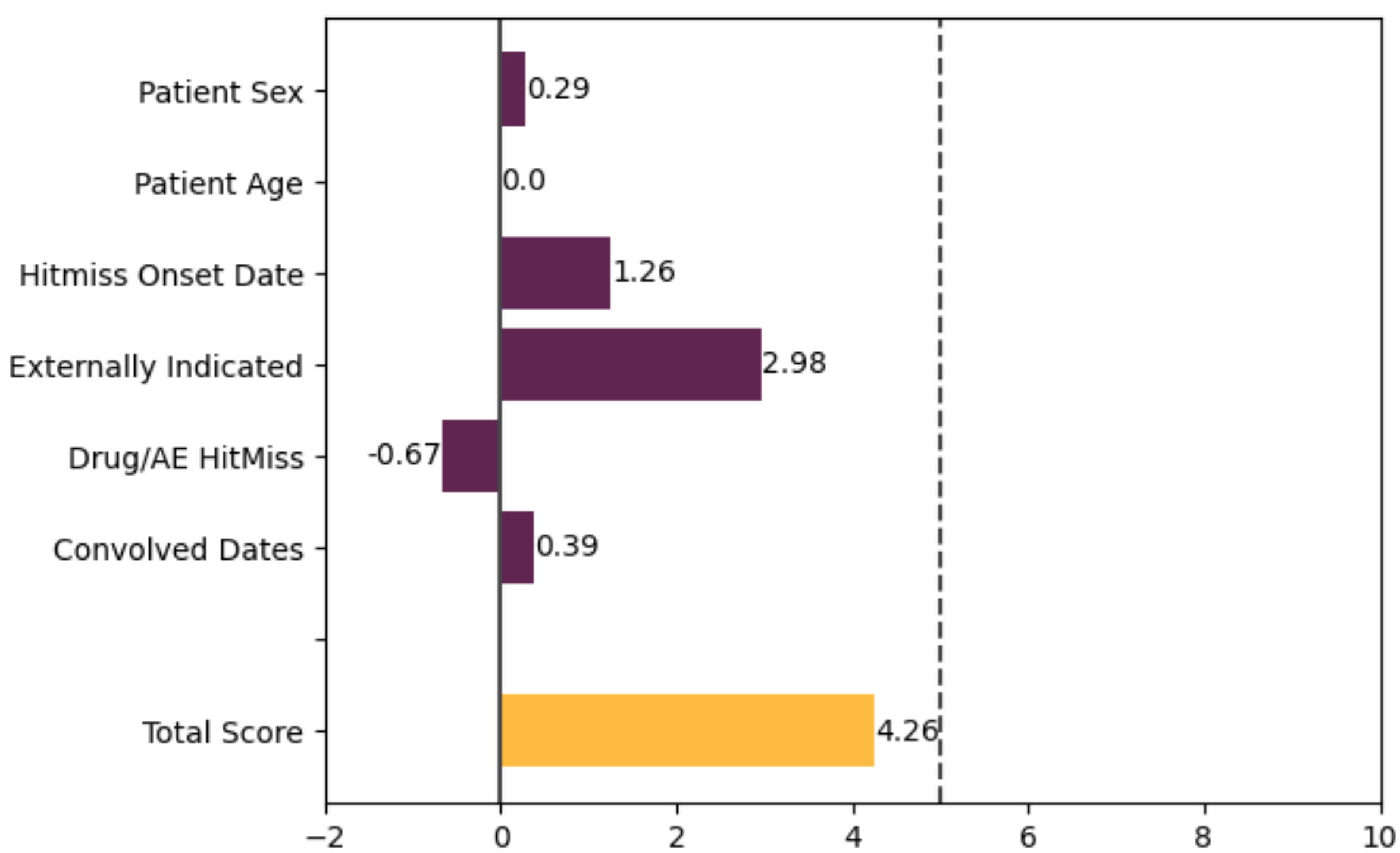